\documentclass[sigconf]{acmart}

\AtBeginDocument{%
  }

\copyrightyear{2026} 
\acmYear{2026}
\setcopyright{cc}
\setcctype{by-nd}

\acmConference[HRI '26]{Special Session on “HRI and Consumer Robots: Building Robots that are Useful and Acceptable”}{March 16--19, 2026}{Edinburgh, Scotland}

\usepackage{tcolorbox}
\usepackage{wrapfig}
\usepackage{graphicx}
\usepackage{array}
\usepackage{ragged2e}

\newcommand{\thumb}[1]{%
  \includegraphics[width=1.28cm,height=1.28cm,keepaspectratio=false,clip]{#1}%
}

\makeatletter
\newlength{\originalauthorbxwd}
\AtBeginDocument{%
  \setlength{\originalauthorbxwd}{\author@bx@wd}
  \setlength{\author@bx@wd}{0.85\originalauthorbxwd}  
}

\renewcommand\@authorfont{\fontsize{10pt}{13pt}\selectfont\sffamily}
\renewcommand\@affiliationfont{\normalsize\normalfont}
\makeatother

\begin{document}

\title{Material Driven HRI Design: Aesthetics as Explainability}

\author{Natalie Friedman}
\affiliation{
  \institution{BTP SAP Innovation}
  \city{Palo Alto}
  \state{CA}
  \country{USA}
}

\author{Kevin Weatherwax}
\affiliation{
  \institution{BTP SAP Innovation}
  \city{Palo Alto}
  \state{CA}
  \country{USA}
}

\author{Chengchao Zhu}
\affiliation{
  \institution{BTP SAP Innovation}
  \city{Palo Alto}
  \state{CA}
  \country{USA}
}

\renewcommand{\shortauthors}{Friedman et al.}

\begin{abstract}
Aesthetics---often treated as secondary to function---guides how people interpret robots’ roles. A great deal of robot designs---both real and fictitious---use sleek industrial aesthetics. These feature hard glossy plastics, hiding as much of the underlying mechanical and electrical components as possible, resembling something akin to a nude humanoid figure. This leaves robots as something of a blank slate to which end-users apply coverings to, often based on media of fiction and non-fiction alike. We argue that designers can take cues from fashion to design interaction and set appropriate expectations. Rather than viewing appearance as decoration, we propose that color, texture, and material choices function as interaction signals. These signals can invite or discourage touch, clarify a robot’s role, and help align user expectations with a robot’s actual capabilities. When done thoughtfully, such cues can create familiarity and legibility; when done poorly, they can lead to wrong expectations. This preliminary paper proposes a framework describing how materials can create explainability by signaling expectations for interaction, task, and environment. We use this framework to do a content analysis of 6 robots.

\end{abstract}

\settopmatter{
  printacmref=false,
  printfolios=false
}
\maketitle
\settopmatter{printacmref=false}
\settopmatter{printfolios=false}

\makeatletter
\renewcommand\maketitle{\@maketitle}
\makeatother

\section{Introduction}

In society, clothing and fashion act as a form of explainability. People routinely infer role, competence, cleanliness, and social norms from colors, textures, and materials. Robotics can leverage this same semiotic system. For example, soft fabrics, like cotton or linen may signal approachability and safety, much as they do in at-home clothing. Conversely, glossy plastics, sharp lines, or industrial metals may communicate precision or restricted interaction, likely not inviting touch. Drawing from fashion, robots can communicate intent through familiar visual, auditory, and tactile affordances. This paper assesses robots materials explainability through assessing what materials are saying about interaction, task, and environment through a content analysis of 6 robots. 

Cultural psychology suggests that perception and action are not purely individual processes but are constituted through engagement with sociomaterial environments. Markus and Kitayama describe the self as developing through ongoing interaction with “ideas, practices, institutions, products, and artifacts,” which together scaffold how people interpret situations and determine appropriate behavior \cite{markus2010cultures}. From this perspective, a robot’s clothing or materiality is not merely aesthetic surface, but part of the cultural context that guides users’ expectations of agency and social roles. In other words, materials help construct the relationship between person and robot before any interaction occurs.

\subsection{Design Research Landscape}
Zimmerman et al. tried to distinguish product design research from RtD (Research Through Design). He explained that design research has an “intention to produce knowledge” rather than informing product development \cite{zimmerman2007research}. The field of fashion, too, has distinguished between the design of products from design to facilitate thinking. More specifically, this appears in mood board types, one of which aims to facilitate thinking (like with RtD), while other mood boards are made as marketing to merely present and communicate products to an audience (like product design research)\cite{cassidy2011mood}.

Cross \cite{cross1982designerly} suggested that design is particularly well suited to ill-defined problem spaces, for which the scientific method is not always the most appropriate approach. Building on this idea, Lupetti et al. \cite{lupetti2021designerly} argued that in HRI, user studies provide only a limited account of the knowledge that design produces. They note that such studies often result in ungeneralizable, stand-alone design implications, in which “very specific robots are produced for very specific problems.” To address this, the authors recommend that HRI designers more clearly articulate the types of contributions their design instances generate, including “concepts [that] stand in between the abstraction of theories and the concreteness of instances.” \cite{lupetti2021designerly, stolterman2010concept}

\subsection{What are "Clothes" for robots?}
Although research on dressing robots remains limited within human–robot interaction, several scholars have explored the potential roles of clothing for robots. Zguda, an HCI researcher, argued that clothing could improve robot approachability, strengthen emotional bonding through customization, and support recognition of characters or roles (e.g., in a restaurant setting) \cite{zguda2023robot}. Similarly, Fitter et al. \cite{fitter2018evaluating} found that people are inclined to personalize the robots they interact with, while Sung et al. \cite{sung2007my, sung2009pimp} showed that people modify their Roombas to make them more noticeable or better aligned with their home’s interior design. Building on this work, Friedman et al. suggest that clothing can also function as protection, enable social signaling, and improve adaptability \cite{friedman2021robots}. For the purposes of this paper, clothes for robots are removable coverings that help a robot fulfill it's social and functional requirements in a given environment. 



\section{Proposed Framework: Materials for Explainability}

This framework outlines how a robot’s material and aesthetic design can communicate meaning before any interaction occurs. We position materials as functional signals that guide how people understand a robot in everyday environments. Specifically, material choices can (1) explain a robot’s task or role, and (2) communicate where the robot belongs, (3) invite or discourage touch interaction. Together, these cues reduce set appropriate expectations for more intuitive human–robot relationships.

\begin{table*}[t]
\centering
\footnotesize
\renewcommand{\arraystretch}{1.0}
\setlength{\tabcolsep}{3pt}
\begin{tabular}{| >{\centering\arraybackslash}m{1.3cm}
                | >{\RaggedRight\arraybackslash}p{2.2cm}
                | >{\RaggedRight\arraybackslash}p{2.6cm}
                | >{\RaggedRight\arraybackslash}p{3.1cm}
                | >{\RaggedRight\arraybackslash}p{2.0cm}
                | >{\RaggedRight\arraybackslash}p{3.6cm} |}
\hline
\textbf{Image} & \textbf{Name} & \textbf{Clothing piece} & \textbf{Task} & \textbf{Setting} & \textbf{Interaction} \\
\hline
\thumb{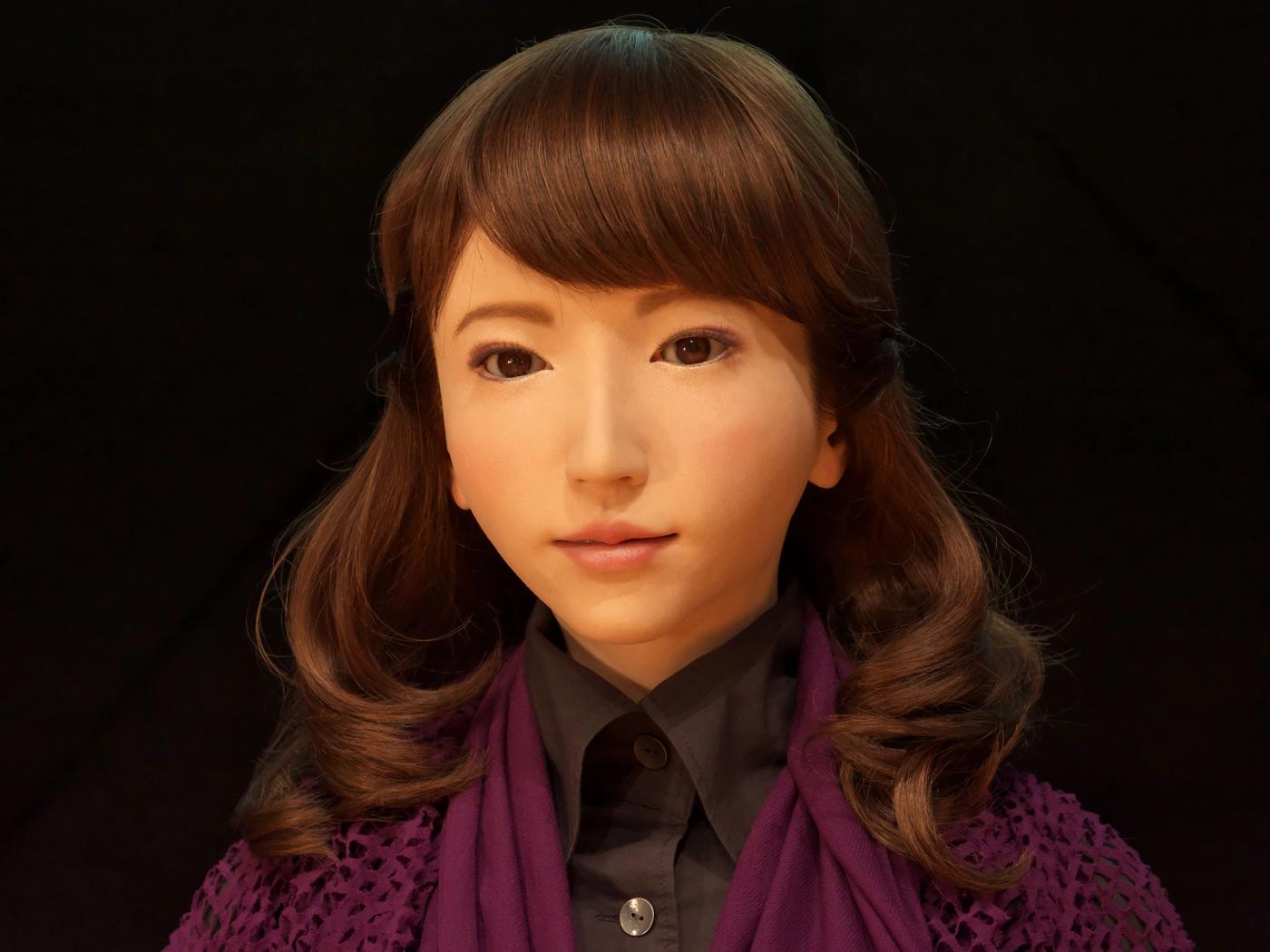} & \centering\textbf{Erica}\arraybackslash & Purple shawl, black button down, and brunette straight hair wig & The formal clothing suggests appearance based tasks, like posing or speaking for a crowd. & Formal clothing suggests Erica is meant to be outside the home, in public. & Because the materials are soft, touch is welcome, but because the materials are higher end, and Erica is a robot, people would touch Erica with care. \\
\hline
\thumb{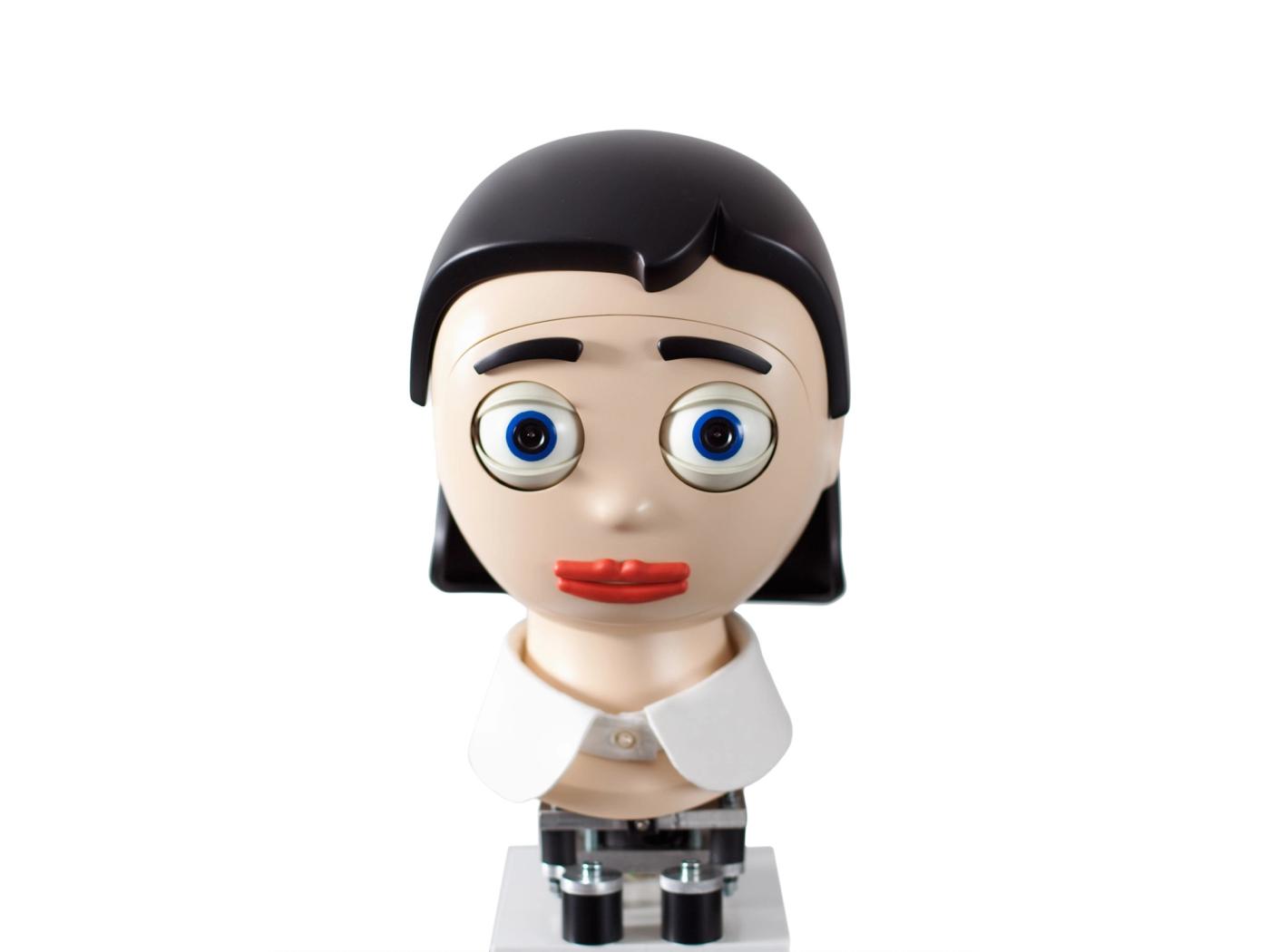} & \centering\textbf{Flobi}\arraybackslash & White collar and removable masks & The collar without a shirt suggests a clean, non-distracting, or neutral identity. However, the robot is meant to imitate people's emotional expressions for kids or people with disabilities. & The collar and simple removable masks suggests this robot belongs at a school. & The texture of the hard plastic does not invite touch, but the toy like shapes and colors do invite touch. The formality of the collar does not necessarily invite touch. \\
\hline
\thumb{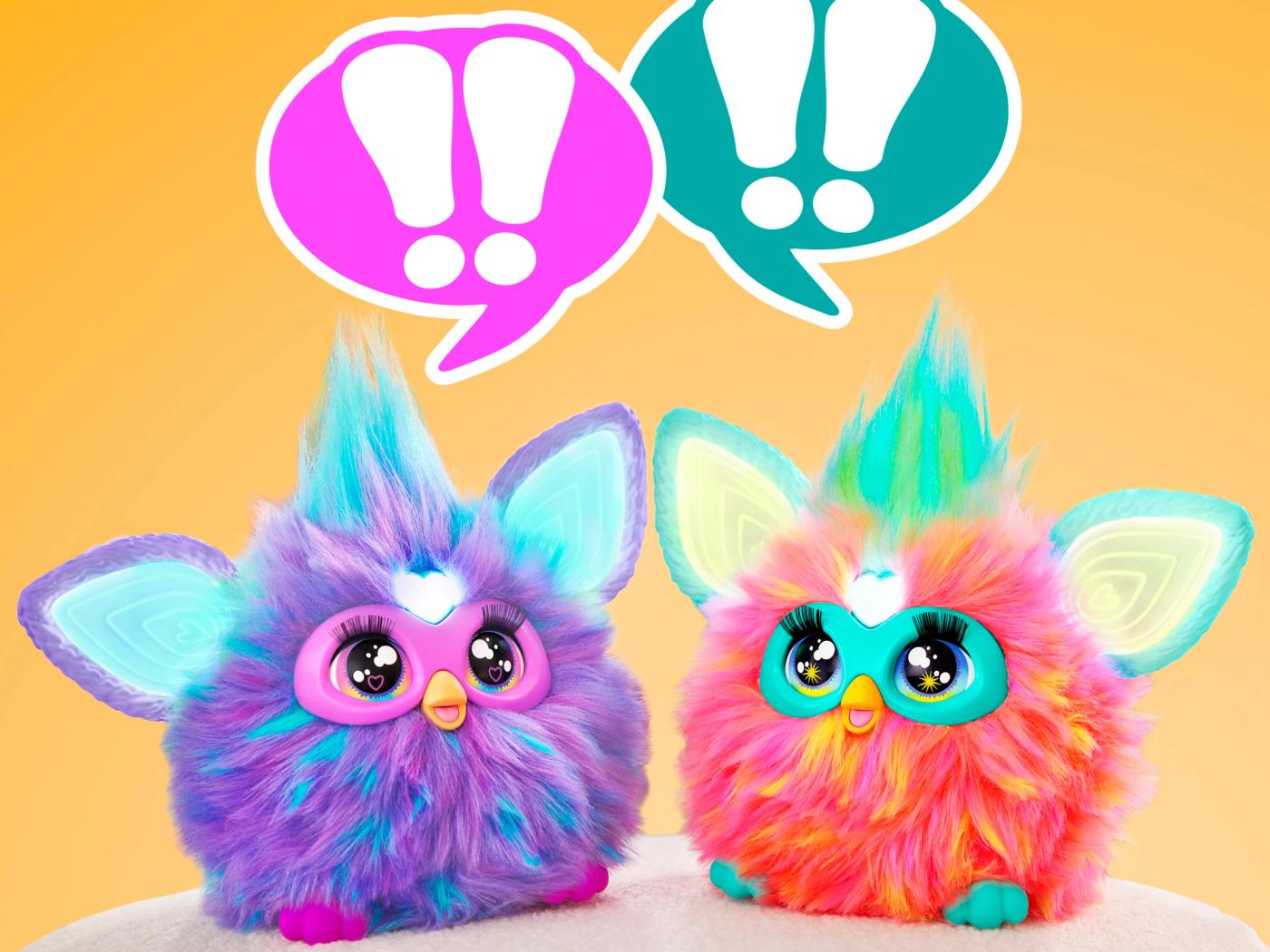} & \centering\textbf{Furby}\arraybackslash & Orange or purple fur & Bright colors invite play-based companionship and entertainment tasks, such as reacting and being cared for like a pet. & Bright colors and soft textures say informal domestic spaces like bedrooms or playrooms where toys belong. & The soft fur strongly invites touch, hugging, and squeezing; interaction is playful, primarily tactile, frequent, and low-risk. \\
\hline
\thumb{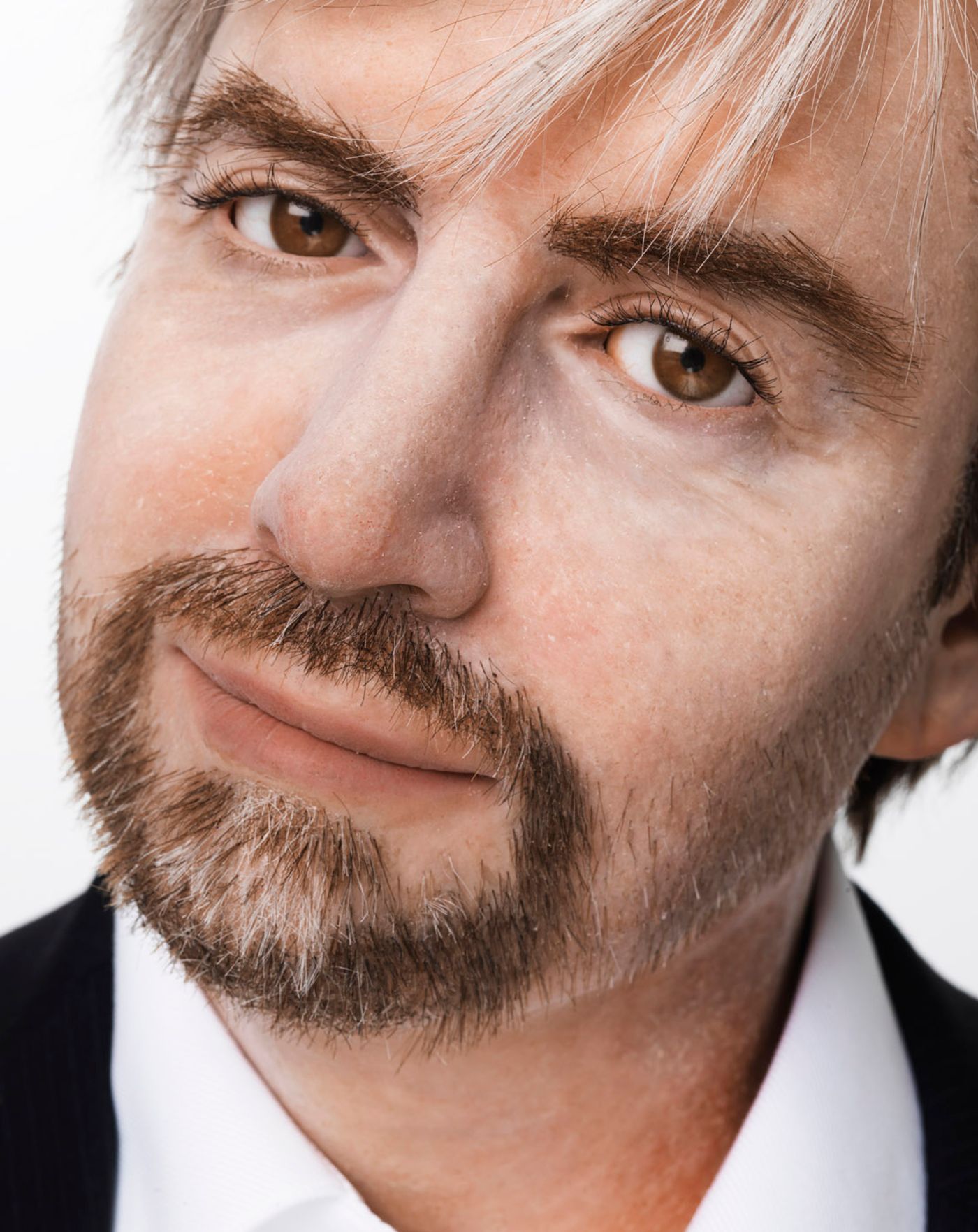} & \centering\textbf{Geminoid DK}\arraybackslash DK & Black and white collar, blonde wig & The formal black color and white shirt suggests lower durability. Conversational-based tasks where the robot stands in for a person, such as, telepresence or public dialogue. & The formality suggests professional or academic environments like labs, offices, or stages. & The realistic clothing and human likeness encourage face-to-face social interaction rather than touch; people are likely to keep physical distance and treat the robot with social etiquette. Similar to Erica, if someone touches Germanoid DK, it will be with care. \\
\hline
\thumb{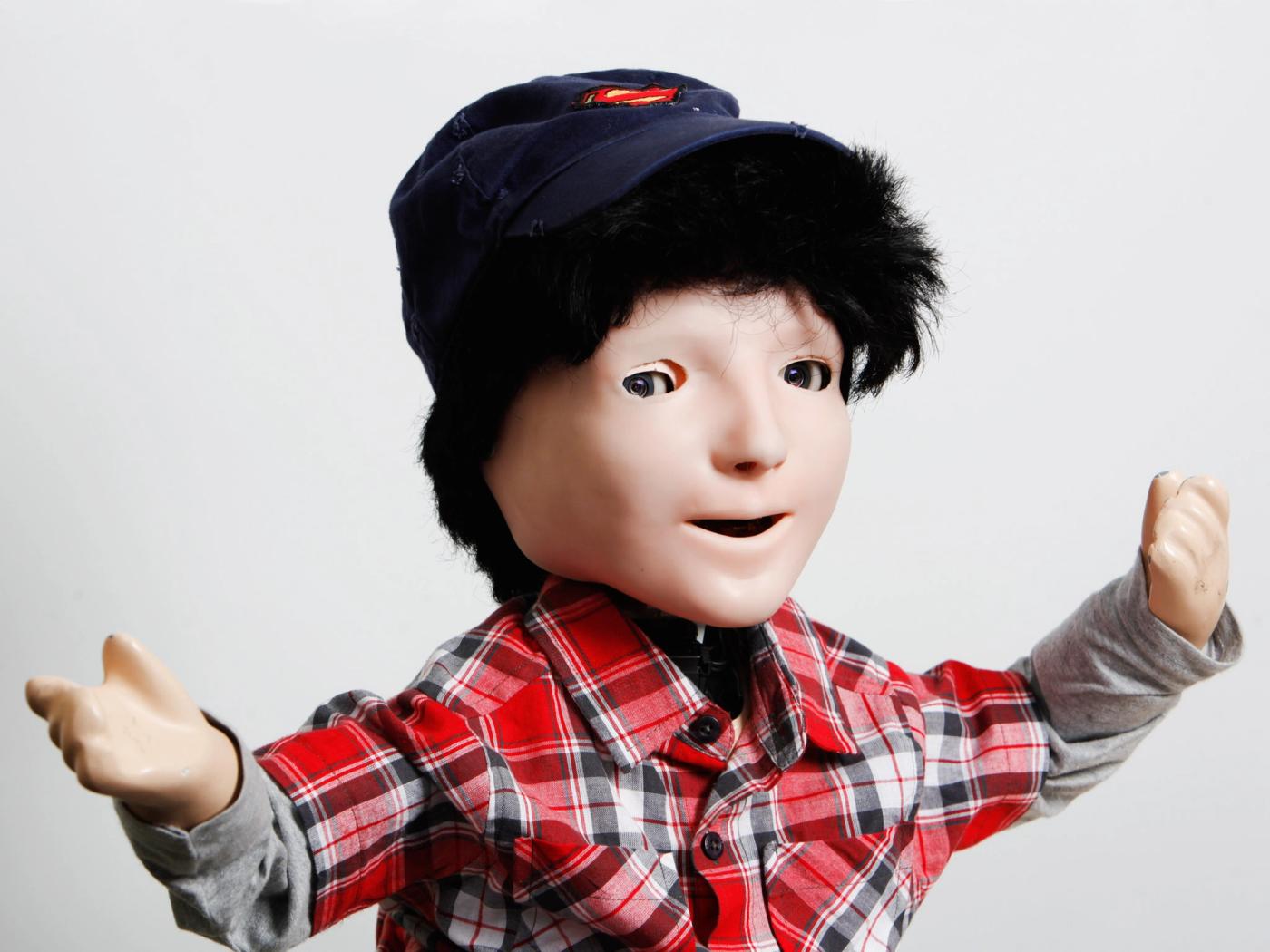} & \centering\textbf{Kaspar}\arraybackslash & Red plaid button-down shirt over a grey long-sleeve t-shirt, superman baseball hat over a short hair black wig & Clothes are similar to the patients' clothes, demonstrating a peer. This could make a patient feel more comfortable doing therapeutic or educational tasks, especially supporting social skills training for children with autism. & The clothes, similar to that of a playful young boy points to settings like classrooms. & The familiar childlike clothing and hat reduce intimidation and encourage gentle touch, play, and frequent physical touch. \\
\hline
\thumb{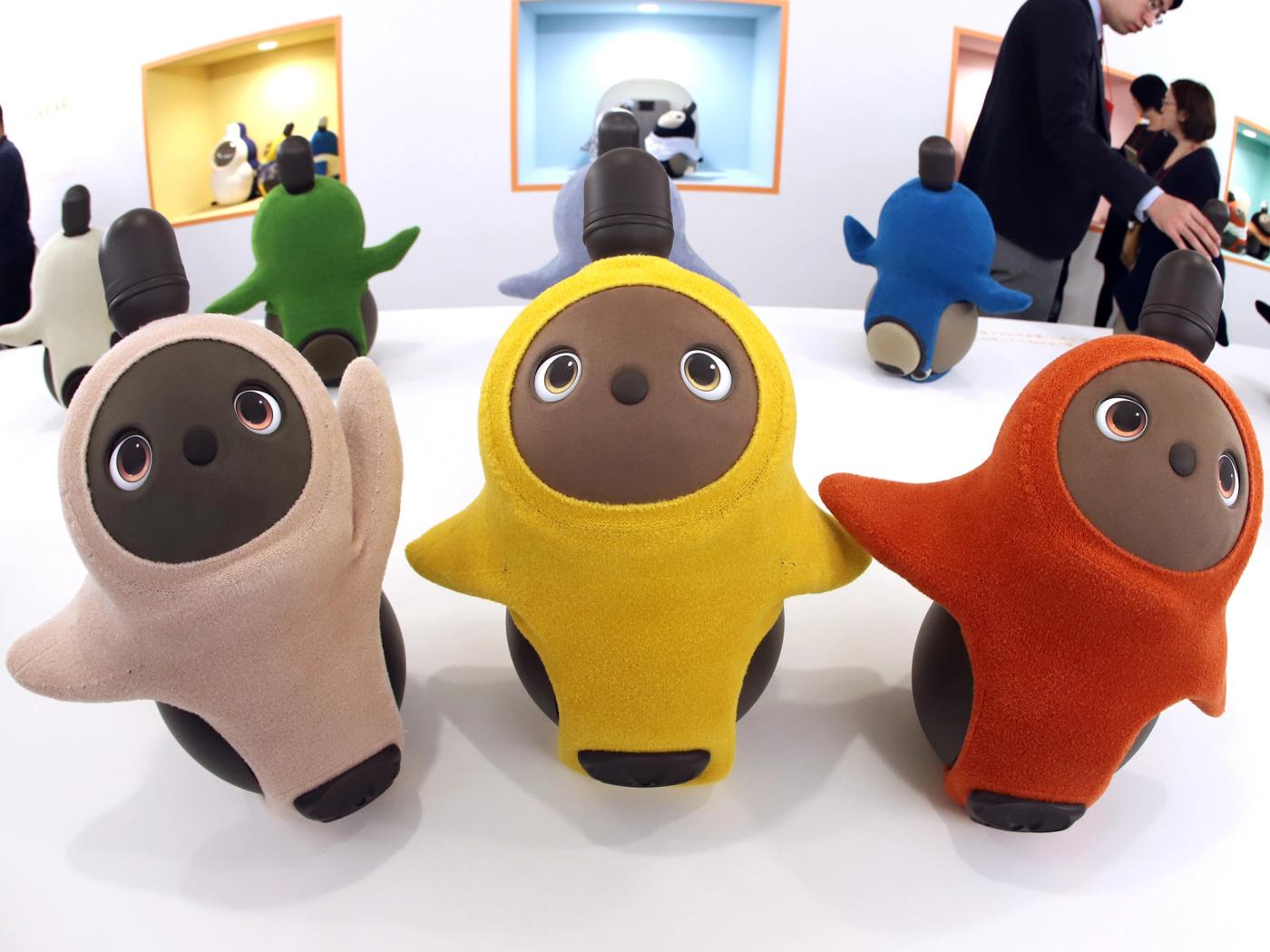} & \centering\textbf{Lovot}\arraybackslash & Pink, yellow, and orange fabric covering "head", "arms", and "stomach." Material appears terry cloth, stretchy, and fitted. & The simple form and playful colors invite emotional companionship and comfort tasks, such as cuddling, following, and responding affectionately to users. & The terry cloth materials point to domestic and caregiving environments like homes or elder care spaces. & The plush, stretchy fabric strongly invites hugging and holding. Interaction is primarily tactile and cozy, similar to interacting with a stuffed animal. \\
\hline
\end{tabular}
\caption{Overview of social robots, their tasks, settings, and interaction expectations. Robots are taken from a ROBOTS list, made by IEEE \cite{robotsguide2026}.}
\label{tab:robot_overview}
\end{table*}
\subsection{Explaining Task}
Clothing has long been used to communicate occupational roles: firefighters’ gear signals danger and authority; chefs’ whites signal cleanliness and food safety; medical professionals’ whites communicates trustworthiness and again, communicates hygiene. Robots can similarly adopt role-appropriate visual languages. For example, a household robot assisting with cooking might use light colors and cleanable textures that evoke kitchen hygiene, while a robot tasked with outdoor maintenance could use durable, weather-resistant materials and high-visibility colors. A service robot in a restaurant might visually align with the brand’s color palette or staff uniforms, reinforcing coherence within the environment. These choices help users quickly understand what the robot is for and how it should be treated. Moreover, this task framing can reduce cognitive load and lower the barrier to acceptance, particularly for first-time users.

\subsection{Explaining Setting Belonging}

Domestic environments are diverse, spanning indoor living spaces with toys sprawled out on the floor, outdoor patios with variable weather, and messy shared kitchens. Materials and aesthetics should reflect these contexts. For outdoor or semi-outdoor use, designers have drawn from materials like Sunbrella, which provide durability, sun and water resistance \cite{friedman2023utility} to protect agriculture robots. In homes with children or older adults, softer materials and muted colors may signal safety and calm. In semi-public domestic spaces, such as restaurants, robots can explain context through color of a waiter's uniform or color of a chef's uniform, distinguishing roles in addition to distinguishing brand. Visual alignment with the surrounding environment can help robots feel less intrusive. Context-aware design requires designers to exist in and reflect on the space they are designing for through understanding real environments and cultural norms.

\subsection{Inviting Interaction}
Material choice strongly shapes whether people feel permitted to touch a robot. Research in HRI suggests that soft, compliant materials invite gentle, exploratory touch, similar to interactions with pets \cite{paterson2023inviting}. Classic studies by Harlow demonstrated that infant rhesus monkeys formed stronger attachment preferences for soft, cloth surrogate mothers over hard, wire ones, even when the latter provided food, highlighting the central role of tactile comfort in attachment formation \cite{harlow1959love}. In domestic settings, where robots may coexist with children, older adults, or pets, these cues matter. A robot designed with plush surfaces may implicitly signal that touch is safe. In contrast, exposed mechanisms or hard shells may discourage physical interaction, which could be appropriate for robots performing hazardous tasks. Researchers should observe interactions in the intended space and design interaction cues to fit the robot’s function and setting norms.





\section{Case study: Robots assessed by aesthetics}

To illustrate how this framework can be applied in practice, we conducted a preliminary small exploratory case study of robots that incorporate clothing. This analysis demonstrates how material choices may communicate task, setting, and interaction expectations.

\subsection{Method}
We propose a framework in which task, setting, and interaction are explainable by clothes. Rather than formal coding, we applied a designerly reading of materials—interpreting how clothing qualities communicated interaction norms, task expectations, and environmental belonging (see Table \ref{tab:robot_overview}). Using this framework, we conducted a preliminary expert, theory-informed assessment of robots listed on the IEEE ROBOTS guide \cite{robotsguide2026}. ROBOTS is an interactive IEEE Spectrum guide that lets people learn about 271 real-world robots through media, technical details, and news. Of the ROBOTS library of 271 robots, which is updated every Friday, only 14 robots had clothes. Because this project is still a work in progress, not all robots with clothes from the ROBOTS list are in this table. Of these 14, we selected six robots that represented a range of materials, roles, and contexts to illustrate diverse applications of clothing.

The analysis was conducted by a researcher with prior experience designing and studying robot clothing in both academic and applied HRI contexts. This expertise enabled nuanced interpretation of material cues, but also introduces subjectivity, which we address as a limitation. As this was an exploratory analysis intended to surface design patterns rather than establish generalizable claims, formal inter-rater reliability was not assessed. Future work will expand the analysis to multiple researchers.


\subsection{Result}
For our results, please refer to Table \ref{tab:robot_overview} which demonstrates a preliminary content analysis of six robots from IEEE's ROBOTS list \cite{robotsguide2026}. Across the six robots analyzed, material cues consistently signaled expectations about touch permissibility. Robots covered in plush or fabric materials (e.g., Furby, Lovot) strongly invited tactile, low-risk interaction, while robots styled in formal clothing (e.g., Erica, Geminoid DK) signaled social interaction over physical touch. In multiple cases, clothing simultaneously communicated task and setting; for example, Kaspar’s childlike attire suggested both therapeutic peer interaction and classroom environments. Formal garments tended to imply public or professional contexts, whereas bright colors and soft textures aligned with domestic and caregiving spaces. Notably, material cues often operated in conjunction with form and naming conventions. These patterns suggest that clothing functions less as decoration and more as a coordination mechanism between task, setting, and interaction norms.

\subsection{Discussion}
The form and the name of the robot, in addition to the clothing, pointed to what types of tasks the robot should do. For example, both the name "Reachy" and the form (unusually long arms) clearly demonstrated that the robot's task was to reach. This points to the fact that so many features (i.e., name, form, color, material) determine our guesses for what a robot could be used for. Moreover, applying the framework revealed that task, setting, and interaction cues are often intertwined rather than separable. Also, in several cases, materials simultaneously signaled social role and interaction permissibility, suggesting that these dimensions may operate as coupled signals rather than independent axes.

Unlike prior work that focuses on form or anthropomorphic features, this analysis highlights materiality as underexplored mechanisms of explainability—providing low-tech, continuous cues that operate before any robot movement or speaking begins.

Given that of the ROBOTS library of 271 robots, only 14 robots had clothes, this tells us that clothes are not initially being designed for most robots, and instead, clothes are being designed long after the design of the robot. 

In the future, it would be helpful to do this content analysis with a wider sample of robots in addition to more researchers. With only one researcher, the subjective nature the content analysis task is enhanced. 

These findings suggest that designers can intentionally use materials to constrain interpretation, not merely enhance likability. Early material decisions could help prevent over-trust, discourage unsafe touch, or align expectations.

The analysis relies on curated images and descriptions rather than observations of real-world interaction, which may differ in practice. Additionally, the ROBOTS database reflects media-visible robots, potentially biasing toward commercial or spectacle-oriented robots.

\section{Risks and Design Responsibilities}

While visual and material cues are powerful, they also carry risk. Using the wrong cues for the wrong context can mislead users or create unsafe situations. For example, overly friendly aesthetics on a powerful factory robot could lead to misuse. A particular challenge lies in expressive elements such as eyes or faces. While eyes can effectively communicate intent and attention, they may also lead users to overestimate perception, understanding, or autonomy. This highlights the importance of designing to set correct expectations, not simply to maximize likability.

Cues may also be interpreted differently across cultural contexts. The same colors or materials can carry conflicting meanings: for instance, red may signal danger or “stop” in some regions but celebration or welcome in others, like China. Such differences can undermine explainability or even create hazardous misunderstandings, especially in safety-critical settings. Designers must therefore consider how aesthetic signals translate across audiences rather than assuming universal interpretations.

An additional risk is gendering robots through clothes. This could reinforce harmful stereotypes, like a female assistant or housekeeper. Given that sound and clothing can be used to reliably establish a robot’s perceived gender \cite{dennler2025designing}, designers must be aware of gender stereotypes in-context. Lastly, we caution against dishonest anthropomorphism \cite{leong2019robot}, in which, for example, a camera should not be placed where eyes are not. Instead, we advocate for appropriate expressiveness: enough to provide legibility, but not so much that it promises features that do or don't exist. Achieving this balance requires careful user research and humility.

\section{Conclusion: Guidance for the Robotics Industry}


If the HRI community were to offer guidance to the robotics industry, it would be to treat aesthetics, materials, and visual language as core interaction design decisions, not afterthoughts. As illustrated in our case study, clothing functioned not merely as decoration but as a mechanism for clarifying task, setting expectations, and shaping interaction boundaries. Drawing from fashion can help robots create clearer first impressions, invite appropriate interactions, and establish more grounded relationships with people in the home. By using familiar cues from clothing specific to interactions, tasks, and contexts, designers can build robots that feel contextually appropriate. In doing so, aesthetics become a form of explainability: robots communicate not only through what they do, but through how they appear.

\bibliographystyle{ACM-Reference-Format}
\bibliography{bibliography}

@article{dennler2025designing,
  title={Designing robot identity: The role of voice, clothing, and task on robot gender perception},
  author={Dennler, Nathaniel and Kian, Mina and Nikolaidis, Stefanos and Matari{\'c}, Maja},
  journal={International Journal of Social Robotics},
  pages={1--22},
  year={2025},
  publisher={Springer}
}

@article{stolterman2010concept,
  title={Concept-driven interaction design research},
  author={Stolterman, Erik and Wiberg, Mikael},
  journal={Human--Computer Interaction},
  volume={25},
  number={2},
  pages={95--118},
  year={2010},
  publisher={Taylor \& Francis}
}

@article{markus2010cultures,
  title={Cultures and selves: A cycle of mutual constitution},
  author={Markus, Hazel Rose and Kitayama, Shinobu},
  journal={Perspectives on psychological science},
  volume={5},
  number={4},
  pages={420--430},
  year={2010},
  publisher={Sage Publications Sage CA: Los Angeles, CA}
}

@inproceedings{zguda2023robot,
  title={Robot in Disguise},
  author={Zguda, Paulina},
  booktitle={International Conference on Human-Computer Interaction},
  pages={268--276},
  year={2023},
  organization={Springer}
}

@inproceedings{sung2009pimp,
  title={" Pimp My Roomba" designing for personalization},
  author={Sung, JaYoung and Grinter, Rebecca E and Christensen, Henrik I},
  booktitle={Proceedings of the SIGCHI Conference on Human Factors in Computing Systems},
  pages={193--196},
  year={2009}
}

@inproceedings{sung2007my,
  title={“My Roomba is Rambo”: intimate home appliances},
  author={Sung, Ja-Young and Guo, Lan and Grinter, Rebecca E and Christensen, Henrik I},
  booktitle={International conference on ubiquitous computing},
  pages={145--162},
  year={2007},
  organization={Springer}
}

@inproceedings{friedman2021robots,
  title={What robots need from clothing},
  author={Friedman, Natalie and Love, Kari and LC, RAY and Sabin, Jenny E and Hoffman, Guy and Ju, Wendy},
  booktitle={Proceedings of the 2021 ACM Designing Interactive Systems Conference},
  pages={1345--1355},
  year={2021}
}

@inproceedings{fitter2018evaluating,
  title={Evaluating the effects of personalized appearance on telepresence robots for education},
  author={Fitter, Naomi T and Chowdhury, Yasmin and Cha, Elizabeth and Takayama, Leila and Matari{\'c}, Maja J},
  booktitle={Companion of the 2018 ACM/IEEE international conference on human-robot interaction},
  pages={109--110},
  year={2018}
}

@inproceedings{lupetti2021designerly,
  title={Designerly ways of knowing in HRI: Broadening the scope of design-oriented HRI through the concept of intermediate-level knowledge},
  author={Lupetti, Maria Luce and Zaga, Cristina and Cila, Nazli},
  booktitle={Proceedings of the 2021 ACM/IEEE International Conference on Human-Robot Interaction},
  pages={389--398},
  year={2021}
}

@article{cross1982designerly,
  title={Designerly ways of knowing},
  author={Cross, Nigel},
  journal={Design studies},
  volume={3},
  number={4},
  pages={221--227},
  year={1982},
  publisher={Elsevier}
}

@article{cassidy2011mood,
  title={The mood board process modeled and understood as a qualitative design research tool},
  author={Cassidy, Tracy},
  journal={Fashion Practice},
  volume={3},
  number={2},
  pages={225--251},
  year={2011},
  publisher={Taylor \& Francis}
}

@inproceedings{zimmerman2007research,
  title={Research through design as a method for interaction design research in HCI},
  author={Zimmerman, John and Forlizzi, Jodi and Evenson, Shelley},
  booktitle={Proceedings of the SIGCHI conference on Human factors in computing systems},
  pages={493--502},
  year={2007}
}

@misc{robotsguide2026,
  title        = {ROBOTS: Your Guide to the World of Robotics},
  howpublished = {\url{https://robotsguide.com/}},
  note         = {Interactive robotics guide created by {\it IEEE Spectrum}, featuring profiles, photos, videos, rankings, and technical information on robots, drones, and autonomous machines},
  year         = {2026},
  organization = {IEEE Spectrum},
}

@article{harlow1959love,
  title={Love in infant monkeys},
  author={Harlow, Harry F},
  journal={Scientific American},
  volume={200},
  number={6},
  pages={68--75},
  year={1959},
  publisher={JSTOR}
}

@inproceedings{leong2019robot,
  title={Robot eyes wide shut: Understanding dishonest anthropomorphism},
  author={Leong, Brenda and Selinger, Evan},
  booktitle={Proceedings of the conference on fairness, accountability, and transparency},
  pages={299--308},
  year={2019}
}

@article{paterson2023inviting,
  title={Inviting robot touch (by design)},
  author={Paterson, Mark},
  journal={ACM Transactions on Human-Robot Interaction},
  volume={12},
  number={2},
  pages={1--17},
  year={2023},
  publisher={ACM New York, NY}
}

@inproceedings{friedman2023utility,
  title={Utility belt for an Agricultural Robot: Reflections on Performing Design Research in the Field},
  author={Friedman, Natalie and Mehta, Asmita and Ju, Wendy},
  booktitle={Companion of the 2023 ACM/IEEE International Conference on Human-Robot Interaction},
  pages={872--874},
  year={2023}
}

\end{document}